\pdfoutput=1

\documentclass[11pt]{article}

\usepackage[]{acl}

\usepackage{times}
\usepackage{latexsym}

\usepackage{amsmath}
\usepackage{makecell}
\usepackage{multirow}
\usepackage[switch]{lineno}  %
\usepackage{url}
\usepackage{footnote}
\usepackage{physics}
\usepackage{amsmath}
\usepackage{tikz}
\usepackage{mathdots}
\usepackage{yhmath}
\usepackage{cancel}
\usepackage{color}
\usepackage{siunitx}
\usepackage{array}
\usepackage{multirow}
\usepackage{amssymb}
\usepackage{gensymb}
\usepackage{tabularx}
\usepackage{booktabs}
\usetikzlibrary{fadings}
\usetikzlibrary{patterns}
\usetikzlibrary{shadows.blur}
\usetikzlibrary{shapes}
\usepackage{colortbl}

\usepackage[T1]{fontenc}

\usepackage[utf8]{inputenc}

\usepackage{microtype}

\newcommand*\samethanks[1][\value{footnote}]{\footnotemark[#1]}

\title{Language-agnostic BERT Sentence Embedding}

\author{Fangxiaoyu Feng\thanks{~~Equal contributions.}, Yinfei Yang\samethanks{\hspace{1.2mm}}\thanks{~~Work done while at Google.}, Daniel Cer, Naveen Arivazhagan, Wei Wang\samethanks \\
Google AI \\
Mountain View \\
  \texttt{\{fangxiaoyu, cer, navari\}@google.com} \\
  \texttt{\{yangyin7, wei.wang.world\}@gmail.com}\\}

\begin{document}
\maketitle
\begin{abstract}
While BERT is an effective method for learning monolingual sentence embeddings for semantic similarity and embedding based transfer learning~\cite{SentenceBERT}, BERT based cross-lingual sentence embeddings have yet to be explored. We systematically investigate methods for learning multilingual sentence embeddings by combining the best methods for learning monolingual and cross-lingual representations including: masked language modeling (MLM), translation language modeling (TLM)~\cite{XLM}, dual encoder translation ranking~\cite{mandy2018}, and additive margin softmax~\cite{yang2019improve}. We show that introducing a pre-trained multilingual language model dramatically reduces the amount of parallel training data required to achieve good performance by 80\%. Composing the best of these methods produces a model that achieves 83.7\% bi-text retrieval accuracy over 112 languages on Tatoeba, well above the 65.5\% achieved by~\citet{laser}, while still performing competitively on monolingual transfer learning benchmarks~\cite{conneau-kiela-2018-senteval}. Parallel data mined from CommonCrawl using our best model is shown to train competitive NMT models for en-zh and en-de. We publicly release our best multilingual sentence embedding model for 109+ languages at \url{https://tfhub.dev/google/LaBSE}.  
\end{abstract}

\section{Introduction}

In this paper, we systematically explore using pretraining language models in combination with the best of existing methods for learning cross-lingual sentence embeddings. Such embeddings are useful for clustering, retrieval, and modular use of text representations for downstream tasks. While existing cross-lingual sentence embedding models incorporate large transformer models, using large pretrained language models is not well explored. Rather in prior work, encoders are trained directly on translation pairs \cite{laser,mandy2018,yang2019improve}, or on translation pairs combined with monolingual input-response prediction \cite{muthu-crosslingual,muse}. 


  
\begin{figure}[!t]
  \centering
  \includegraphics[width=.96\linewidth]{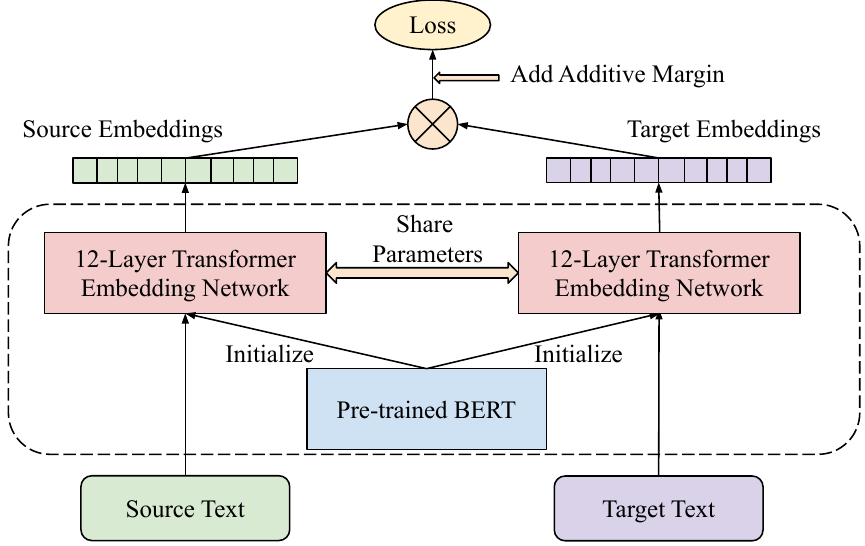}
  \caption{Dual encoder model with BERT based encoding modules.}
  \label{fig:dual_encoder}
\end{figure}

In our exploration, as illustrated in figure \ref{fig:dual_encoder}, we make use of dual-encoder models, which have been demonstrated as an effective approach for learning bilingual sentence embeddings~\cite{mandy2018,yang2019improve}. However, diverging from prior work, rather than training encoders from scratch, we investigate using pre-trained encoders based on large language models.
We contrast models with and without additive margin softmax~\cite{yang2019improve}\footnote{We also investigate the impact of mining hard negatives~\cite{mandy2018}, but found it doesn't provide additional gain on top of other approaches. See supplemental material for details.}. Figure~\ref{fig:work_place} illustrates where our work stands (shaded) in the field of LM pre-training and sentence embedding learning.


Our massively multilingual models outperform the previous state-of-the-art on large bi-text retrieval tasks including the United Nations (UN) corpus ~\cite{uncorpus} and BUCC~\cite{bucc2018}.
Table \ref{tab:model} compares our best model with other recent multilingual work.

Both the UN corpus and BUCC cover resource rich languages (fr, de, es, ru, and zh). We further evaluate our models on the Tatoeba retrieval task~\cite{laser} that covers 112 languages.
Compare to LASER~\cite{laser}, our models perform significantly better on low-resource languages, boosting the overall accuracy on 112 languages to 83.7\%, from the 65.5\% achieved by the previous state-of-art.
Surprisingly, we observe our models performs well on 30+ Tatoeba languages for which we have no explicit monolingual or bilingual training data. 
Finally, our embeddings perform competitively on the SentEval sentence embedding transfer learning benchmark~\cite{conneau-kiela-2018-senteval}.

The contributions of this paper are:
\begin{itemize}
\item A novel combination of pre-training and dual-encoder finetuning to boost translation ranking performance, achieving a new state-of-the-art on bi-text mining.
\item A publicly released multilingual sentence embedding model \emph{spanning 109+ languages}.
\item Thorough experiments and ablation studies to understand the impact of pre-training, negative sampling strategies, vocabulary choice, data quality, and data quantity.
\end{itemize}
We release the pre-trained model at \url{https://tfhub.dev/google/LaBSE}.


\begin{figure}[!t]
  \centering
  \includegraphics[width=\linewidth]{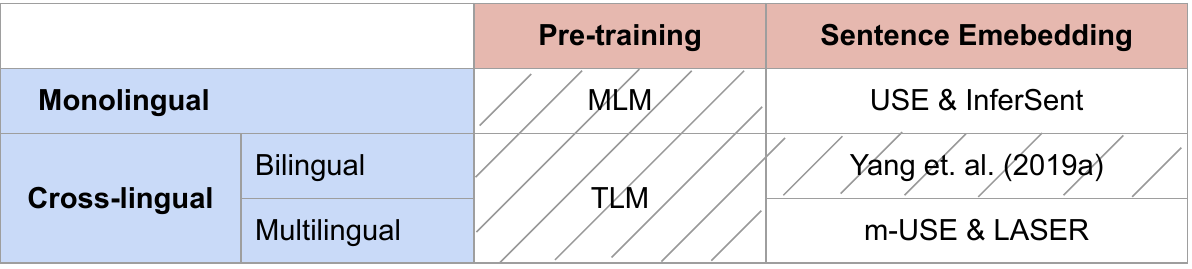}
  \caption{Where our work stands (shaded) vs. related work in LM pre-training and sentence embedding learning.}
  \label{fig:work_place}
\end{figure}

\begin{table}[!t]
    \centering
    \resizebox{\linewidth}{!}{
    \begin{tabular}{l | rrrrr }
        \textbf{Model} & \textbf{Langs} & \textbf{Model} & \textbf{HN} & \textbf{AMS} & \textbf{Pre-train} \\ \hline
         LASER                   &  97 & seq2seq     & N/A & N/A & N \\
         \citet{yang2019improve} &   2 & DE      & Y & Y & N \\
         m-USE                   &  16 & DE      & Y & Y & N \\
         LaBSE                   & 109 & DE      & N & Y & Y \\
    \end{tabular}
    }
    \caption{LaBSE model compared to other recent cross-lingual embedding models. \textbf{[DE]}: Dual Encoder. \textbf{[HN]}: Hard Negative. \textbf{[AMS]}: Additive Margin Softmax. \textbf{[PT]}: Pre-training.}
    \label{tab:model}
\end{table}

\section{Cross-lingual Sentence Embeddings}
\label{sec:techs}

Dual encoder models are an effective approach for learning cross-lingual embeddings~\cite{mandy2018,yang2019improve}.
Such models consist of paired encoding models that feed a scoring function. The source and target sentences are encoded separately. Sentence embeddings are extracted from each encoder. Cross-lingual embeddings are trained using a translation ranking task with in-batch negative sampling: 

\begin{equation}
\label{eq:softmax_ams}
\mathcal{L} = -\frac{1}{N} {\sum_{i=1}^{N}log\frac{e^{\phi(x_i, y_i)}} { e^{\phi(x_i, y_i)} + \sum_{n=1, n \neq i}^{N} e^{\phi(x_i, y_n)}}}
\end{equation}

The embedding space similarity of $x$ and $y$ is given by $\phi(x, y)$, typically $\phi(x, y)=x  y^T$. The loss attempts to rank $y_i$, the true translation of $x_i$, over all $N-1$ alternatives in the same batch. Notice that $\mathcal{L}$ is asymmetric and depends on whether the softmax is over the source or the target sentences. For bidirectional symmetry, the final loss can sum the source-to-target, $\mathcal{L}$, and target-to-source, $\mathcal{L}'$, losses~\cite{yang2019improve}:
\begin{equation}
    \bar{\mathcal{L}} = \mathcal{L} + \mathcal{L}'
\label{eq:softmax_ams_combined}
\end{equation}

Dual encoder models trained using a translation ranking loss directly maximize the similarity of translation pairs in a shared embedding space. 

\subsection{Additive Margin Softmax}

Additive margin softmax extends the scoring function $\phi$ by introducing margin $m$ around positive pairs~\cite{yang2019improve}:

\begin{equation}
    \label{eq:scoring_am}
    \phi'(x_i, y_j) =
        \begin{cases}
          \phi(x_i, y_j) - m    & \quad \text{if } i = j \\
          \phi(x_i, y_j)        & \quad \text{if } i \neq j
        \end{cases}
\end{equation}

The margin, $m$, improves the separation between translations and nearby non-translations. Using $\phi'(x_i, y_j)$ with the bidirectional loss $\bar{\mathcal{L}}_s$, we obtain the additive margin loss

\begin{equation}
\label{eq:softmax_ams}
\mathcal{L} = -\frac{1}{N} {\sum_{i=1}^{N}\frac{e^{\phi(x_i, y_i) - m}} { e^{\phi(x_i, y_i) - m} + \sum_{n=1, n \neq i}^{N} e^{\phi(x_i, y_n)}}}
\end{equation}

\subsection{MLM and TLM Pre-training}
Only limited prior work has combined dual encoders trained with a translation ranking loss with encoders initialized using large pre-trained language models~\cite{ziyi2021}. We contrast using a randomly initialized transformer, as was done in prior work ~\cite{mandy2018,yang2019improve}, with using a large pre-trained language model. For pre-training, we combined Masked language modeling (MLM)~\cite{BERT} and Translation language modeling (TLM)~\cite{XLM}. MLM is a variant of a cloze task, whereby a model uses context words surrounding a [MASK] token to try to predict what the [MASK] word should be. TLM extends this to the multilingual setting by modifying MLM training to include concatenated translation pairs.

Multilingual pre-trained models such as mBERT~\cite{BERT}, XLM~\cite{XLM} and XLM-R~\cite{XLMR} have led to exceptional gains across a variety of cross-lingual natural language processing tasks~\cite{hu2020xtreme}. 
However, without a sentence level objective, they do not directly produce good sentence embeddings.
As shown in \citet{hu2020xtreme}, the performance of such models on bitext retrieval tasks is very weak, e.g  XLM-R Large gets 57.3\% accuracy on a selected 37 languages\footnote{The number is counted from official evaluation script despite the original paper says 33 languages.} from the Tatoeba dataset compared to 84.4\% using LASER (see performance of more models in table \ref{tab:tatoeba}).
We contribute a detailed exploration that uses pre-trained language models to produce useful multilingual sentence embeddings.

\begin{figure*}[!t]
  \centering
  \includegraphics[width=\linewidth]{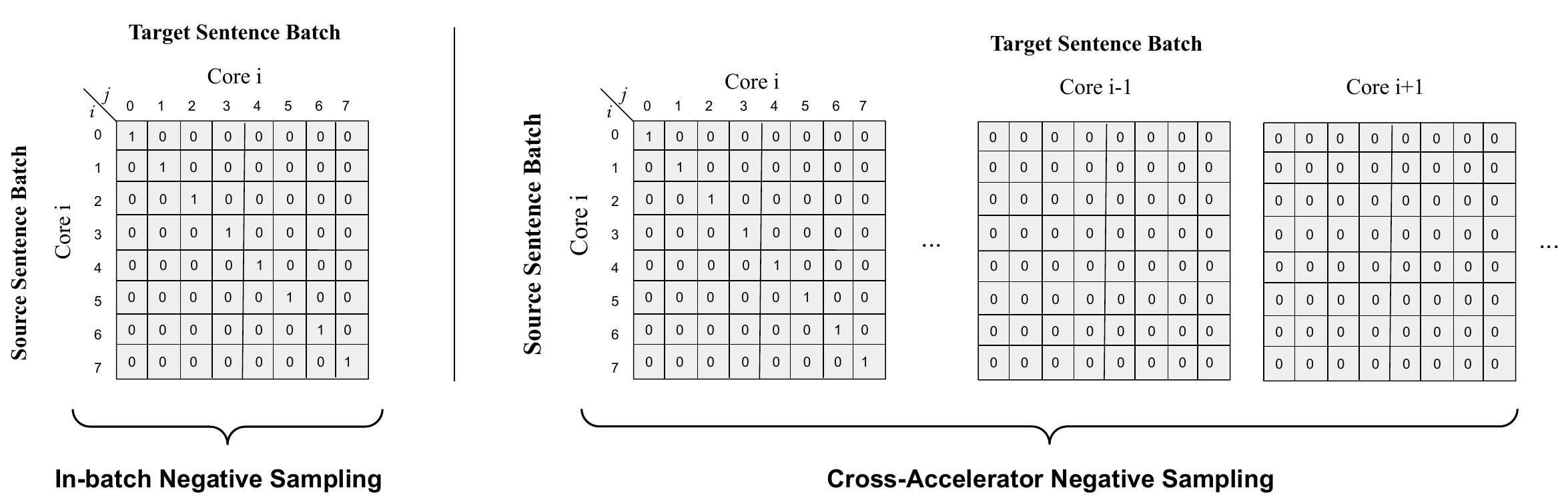}
  \caption{Negative sampling example in a dual encoder framework.  \textbf{[Left]}: The in-batch negative sampling in a single core; \textbf{[Right]}: \emph{Synchronized multi-accelerator negative sampling} using n TPU cores and batch size 8 per core with examples from other cores are all treated as negatives.}
  \label{fig:cross_tpu}
\end{figure*}

\section{Corpus and Training Details}

\subsection{Corpus}
\label{sec:corpus}

We use bilingual translation pairs and monolingual data in our experiments\footnote{See the detailed list of supported languages in supplemental material. }.

\paragraph{Monolingual Data}
\label{sec:monolingual_data}
We collect monolingual data from CommonCrawl\footnote{\url{https://commoncrawl.org/}} and Wikipedia\footnote{\url{https://www.wikipedia.org/}}.
We use the 2019-35 version of CommonCrawl with heuristics from \citet{Raffel2019ExploringTL} to remove noisy text.
Additionally, we remove short lines $<10$ characters and those $>5000$ characters.\footnote{Long lines are usually JavaScript or attempts at SEO.}
The wiki data is extracted from the 05-21-2020 dump using WikiExtractor\footnote{\url{https://github.com/attardi/wikiextractor}}. An in-house tool splits the text into sentences.
The sentences are filtered using a sentence quality classifier.\footnote{
The quality classifier is trained using sentences from the main content of webpages as positives and text from other areas as negatives.}
After filtering,  we obtain 17B monolingual sentences, about 50\% of the unfiltered version.
The monolingual data is only used in customized pre-training.

\paragraph{Bilingual Translation Pairs}
The translation corpus is constructed from web pages using a bitext mining system similar to the approach described in \citet{jakob2010}.
The extracted sentence pairs are filtered by a pre-trained contrastive-data-selection~(CDS) scoring model~\cite{wei2018}.
Human annotators manually evaluate sentence pairs from a small subset of the harvested pairs and mark the pairs as either GOOD or BAD translations.
The data-selection scoring model threshold is chosen such that 80\% of the retained pairs from the manual evaluation are rated as GOOD.
We further limit the maximum number of sentence pairs to 100 million for each language to balance the data distribution. Many languages still have far fewer than 100M sentences. The final corpus contains  6B translation pairs.\footnote{Experiments in later sections show that even 200M pairs across all languages is sufficient.}
The translation corpus is used for both dual encoder training and customized pre-training.

\subsection{Configurations}
In this section, we describe the training details for the dual encoder model.
A transformer encoder is used in all experiments~\cite{vaswani2017attention}.
We train two versions of the model, one uses the public BERT multilingual cased vocab with vocab size 119,547 and a second incorporates a customized vocab extracted over our training data.
For the customized vocab, we employ a wordpiece tokenizer~\cite{sennrich-etal-2016-neural}, with a cased vocabulary extracted from the training set using TF Text.\footnote{\url{https://github.com/tensorflow/text}}
The language smoothing exponent for the vocab generation tool is set to 0.3 to counter imbalances in the amount of data available per language.
The final vocabulary size is 501,153.

The encoder architecture follows the BERT Base model, with 12 transformer blocks, 12 attention heads and 768 per-position hidden units.
The encoder parameters are shared for all languages.
Sentence embeddings are extracted as the $l_2$ normalized {\tt [CLS]} token representations from the last transformer block.\footnote{During training, the sentence embeddings after normalization are multiplied by a scaling factor. Following \citet{mutty2018}, we set the scaling factor to 10. We observe that the scaling factor is important for training a dual encoder model with the normalized embeddings.} 

Our models are trained on Cloud TPU V3 with 32-cores using a global batch size of 4096 with a max sequence length of 128, using the AdamW~\cite{loshchilov2018decoupled} optimizer with initial learning rate 1e-3, and linear weight decay.
We train for 50k steps for models with pre-training, and 500k steps for models without pre-training.
We observe that even further training did not change the performance significantly.
The default margin value for additive margin softmax is set to 0.3.
Hyperparameters are tuned on a held-out development set.

\subsection{Cross-Accelerator Negative Sampling}
\label{sec:tpu}

Cross-lingual embedding models trained with in-batch negative samples benefit from large training batch sizes~\cite{mandy2018}.
Resource intensive models like BERT, are limited to small batch sizes due to memory constraints. While data-parallelism does allow us to increase the global batch size by using multiple accelerators, the batch-size on an individual cores remains small. For example, a 4096 batch run across 32 cores results in a local batch size of 128, with each example then only receiving 127 negatives. 

We introduce \emph{cross-accelerator negative sampling}, which is illustrated in figure \ref{fig:cross_tpu}.\footnote{While our experiments use TPU accelerators, the same strategy can also be applied to models trained on GPU.}
Under this strategy each core encodes its assigned sentences and then the encoded sentence representations from all cores are broadcast as negatives to the other cores.
This allows us to fully realize the benefits of larger batch sizes while still distributing the computationally intensive encoding work across multiple cores.

Note the dot-product scoring function makes it efficient to compute the pairwise scores in the same batch with matrix multiplication. In figure \ref{fig:cross_tpu}, the value in the grids indicates the ground truth labels, with all positive labels located in diagonal grids.
A softmax function is applied on each row.

\subsection{Pre-training}
The encoder is pre-trained with Masked Language Model~(MLM)~\cite{BERT} and Translation Language Model~(TLM)~\cite{XLM}\footnote{Diverging from \citet{XLM}, we do not provide a language hint to encourage multilinguality.
} training on the monolingual data and bilingual translation pairs, respectively.
For an $L$ layer transformer encoder, we train using a 3 stage progressive stacking algorithm~\cite{pbert}, where we first learn a $\frac{L}{4}$ layers model and then $\frac{L}{2}$ layers and finally all $L$ layers.
The parameters of the models learned in the earlier stages are copied to the models for the subsequent stages. 

Pre-training uses TPUv3 with 512-cores and a batch size of 8192. The max sequence length is set to 512 and 20\% of tokens (or 80 tokens at most) per sequence are masked for MLM and TLM predictions. For the three stages of progressive stacking, we respectively train for 400k, 800k, and 1.8M steps using all monolingual and bilingual data.

\section{Evaluation Tasks}

\subsection{Bitext Retrieval}
We evaluate models on three bitext retrieval tasks: United Nations~(UN), Tatoeba, and BUCC.
All tasks are to retrieve the correct English translation for each non-English sentence. 

\paragraph{United Nations~(UN)} contains 86,000 sentence aligned bilingual documents over five language pairs: en-fr, en-es, en-ru, en-ar and en-zh~\cite{uncorpus}. 
A total of 11.3 million\footnote{About 9.5 million after de-duping.} aligned sentence pairs can be extract from the document pairs. The large pool of translation candidates makes this data set particularly challenging.

\paragraph{Tatoeba} evaluates translation retrieval over 112 languages~\cite{laser}. The dataset contains up to 1,000 sentences per language along with their English translations. 
We evaluate performance on the original version covering all 112 languages, and also the 36 languages version from the XTREME benchmark~\cite{hu2020xtreme}.

\paragraph{BUCC} is a parallel sentence mining shared task~\cite{bucc2018}.
We use the 2018 shared task data, containing four language pairs: fr-en, de-en, ru-en and zh-en.
For each pair, the task provides monolingual corpora and gold true translation pairs.
The task is to extract translation pairs from the monolingual data, which are evaluated against the ground truth using F1. 
Since the ground truth for the BUCC test data is not released, we follow prior work using the BUCC training set for evaluation rather than training~\cite{muse,hu2020xtreme}. Sentence embedding cosine similarity is used to identify the translation pairs.\footnote{Reranking models can further improve performance (e.g. margin based scorer~\cite{artetxe2018a} and BERT based classifier~\cite{yang2019improve}). However, this ss tangential to assessing the raw embedding retrieval performance.}

\begin{table*}[!t]
    \centering
    \small
    \begin{tabular}{l  |@{\hskip 3mm} r r r r r @{\hskip 6mm} r r}
         \multirow{2}{*}{\textbf{Model}}
         & \multicolumn{5}{c}{\bf UN (en~$\rightarrow$~xx)} &  \multicolumn{2}{c}{\bf Taoeba (xx~$\rightarrow$~en)} \\
         &  \textbf{es} & \textbf{fr} & \textbf{ru} & \textbf{zh} & \textbf{avg} & \textbf{36 Langs} & \textbf{All Langs}\\ \hline
         LASER~\cite{laser}              & --   & --   & --   & --  & --  & 84.4 & 65.5 \\
         \emph{m}-USE~\cite{muse}        & 86.1 & 83.3 & 88.9 & 78.8 & 84.3 & --   & --   \\  
         \citet{yang2019improve}         & 89.0 & 86.1 & 89.2 & \textbf{87.9} & 88.1 &--   & -- \\  \hline 
         \rule{-2pt}{8pt}
         Base w/ mBERT Vocab              & 67.7	& 57.0 & 70.2 & 71.9 & 66.7 & 92.8 & 79.1\\
         ~~~~ + PT                 & 68.5 & 59.8 & 65.8 & 71.7 & 66,5 & 92.7 & 78.6 \\
         ~~~~ + AMS                & 88.2 & 84.5 & 88.6 & 86.4 & 86.9 & 93.7 & 81.2 \\
         ~~~~ + AMS + PT           & 89.3 & 85.7 & 89.3 & 87.2 & 87.9 & 93.2 & 78.4 \\
         Base w/ Customized Vocab   & & & & & & &          \\
         ~~~~ + AMS                           & 90.6 & 86.5 & 89.5 & 86.8 & 88.4 & 94.8 & 82.6\\
         ~~~~ + AMS + PT (\textbf{LaBSE})     & \textbf{91.1} & \textbf{88.3} & \textbf{90.8} & 87.7 & \textbf{89.5} & \textbf{95.0} & \textbf{83.7} \\
    \end{tabular}
    \caption{UN (P@1) and Taoteba (Average accuracy) performance for different model configurations. \textbf{Base} uses a bidirectional dual encoder model. \textbf{[AMS]}: Additive Margin Softmax. \textbf{[PT]}: Pre-training.}
    \label{tab:br_result}
\end{table*}

\subsection{Downstream Classification}

We also evaluate the transfer performance of multilingual sentence embeddings on downstream classification tasks from the SentEval benchmark~\cite{conneau-kiela-2018-senteval}.
We evaluate on select tasks from SentEval including: (\textbf{MR}) movie reviews~\cite{mr}), (\textbf{SST}) sentiment analysis~\cite{sst}, (\textbf{TREC}) question-type~\cite{trec}, (\textbf{CR}) product reviews~\cite{cr}, (\textbf{SUBJ}) subjectivity/objectivity~\cite{subj}, (\textbf{MPQA}) opinion  polarity~\cite{mpqa}, and (\textbf{MRPC}) paraphrasing detection~\cite{mrpc}. While SentEval is English only, we make use of this benchmark in order to directly compare to prior work on sentence embedding models.

\section{Results}

Table \ref{tab:br_result} shows the performance on the UN and Tatoeba bitext retrieval tasks and compares against the prior state-of-the-art bilingual models \citet{yang2019improve}, LASER~\cite{laser}, and the multilingual universal sentence encoder~(\emph{m}-USE)~\cite{muse}\footnote{universal-sentence-encoder-multilingual-large/3}.
Row 1-3 show the performance of baseline models, as reported in the original papers.

Row 4-7 shows the performance of models that use the public mBERT vocabulary. 
The baseline model shows reasonable performance on UN ranging from 57\%-71\% P@1.
It also perform well on Tatoeba with 92.8\% and 79.1\% accuracy for the 36 language group and all languages, respectively.
Adding pre-training both helps models converge faster (see details in section \ref{sec:effec_pt}) and improves performance on the UN retrieval using both vocabularies. Pre-training also helps on Taoeba, but only using the customized vocabulary.\footnote{The coverage of the public mBERT vocabulary on the tail languages are bad with many [UNK] tokens in those languages, e.g. the [UNK] token rate is 71\% for language \textbf{si}, which could be reason the pre-training doesn't help on the tatoeba task.}
Additive margin softmax significantly improves the performance on all model variations.

The last two rows contain two models using the customized vocab. Both of them are trained with additive margin softmax given the strong evidence from the experiments above.
Both models outperform the mBERT vocabulary based models, and the model with pre-training performs best of all.
The top model~(Base w/ Customized Vocab + AMS + PT) achieves a new state-of-the-art on 3 of the 4 languages, with P@1 91.1, 88.3, 90.8 for en-es, en-fr, en-ru respectively. 
It reaches 87.7 on zh-en, only 0.2 lower than the best bilingual en-zh model and \emph{nearly 9 points better than the previous best multilingual model}. 
On Tatoeba, the best model also outperform the baseline model by a large margin, with +10.6 accuracy on the 36 language group from XTREME and +18.2 on all languages.

It is worth noting that all our models perform similarly on Tatoeba but not on UN. This suggests it is necessary to evaluate on large scale bitext retrieval tasks to better discern differences between competing models. In the rest of the paper we refer to \textbf{LaBSE} as the best performing model here \textit{Base w/ Customized Vocab + AMS + PT}, unless otherwise specified.

\begin{table*}[!htb]
    \centering
    \resizebox{\textwidth}{!}{
    \begin{tabular}{l | l | r r r |  r r r |  r r r  | r r r }
        \hline
        \multirow{2}{*}{\bf} & \multirow{2}{*}{\bf Models} &\multicolumn{3}{c|}{\bf fr-en} & \multicolumn{3}{c|}{\bf de-en} & \multicolumn{3}{c|}{\bf ru-en} & \multicolumn{3}{c}{\bf zh-en} \\
        \cline{3-14}
        & &  P & R & F &  P & R & F &  P & R & F &  P & R & F\\ \hline 
        \rule{-2pt}{12pt}
        \multirow{3}{*}{\rotatebox{90}{Forward}}
        & \citet{artetxe2018a}      & 82.1 & 74.2 & 78.0 & 78.9 & 75.1 & 77.0 & - & - & - & - & - & - \\
        & \citet{yang2019improve}   & \textbf{86.7} & 85.6 & 86.1 & 90.3 & 88.0 & 89.2 & 84.6 & 91.1 & 87.7 &  86.7 & \textbf{90.9} & 88.8 \\ 
        & LaBSE                     & 86.6 & \textbf{90.9} & \textbf{88.7} & \textbf{92.3} & \textbf{92.7} & \textbf{92.5} & \textbf{86.1} & \textbf{91.9} & \textbf{88.9} & \textbf{88.2} & 89.7 & \textbf{88.9} \\  \hline
        \rule{-2pt}{12pt}
        \multirow{3}{*}{\rotatebox{90}{Backward}}
        & \citet{artetxe2018a}      & 77.2 & 72.7 & 74.7 & 79.0 & 73.1 & 75.9 & - & - & - & - & - & - \\ 
        & \citet{yang2019improve}   & 83.8 & 85.5 & 84.6 & 89.3 & 87.7 & 88.5 & 83.6 & 90.5 & 86.9 & \textbf{88.7} & 87.5 & 88.1\\
        & LaBSE                     & \textbf{87.1} &\textbf{ 88.4} & \textbf{87.8} & \textbf{91.3} & \textbf{92.7} & \textbf{92.0} & \textbf{86.3} & \textbf{90.7} & \textbf{88.4} & 87.8 & \textbf{90.3} & \textbf{89.0} \\ 
    \end{tabular}}
    \caption{[P]recision, [R]ecall and [F]-score of BUCC training set score with cosine similarity scores. The thresholds are chosen for the best F scores on the training set. Following the naming of BUCC task~\protect\cite{bucc2018}, we treat en as the target and the other language as source in forward search. Backward is vice versa.}
    \label{tab:bucc_train}
\end{table*}

Table \ref{tab:bucc_train} provides LaBSE's retrieval performance on BUCC, comparing against strong baselines from \citet{artetxe2018a} and \citet{yang2019improve}.
Following prior work, we perform both forward and backward retrieval. Forward retrieval treats en as the target and the other language as the source, and backward retrieval is vice versa. LaBSE not only systematically outperforms prior work but also covers all languages within a single model. The previous state-of-the-art required four separate bilingual models~\cite{yang2019improve}.

\subsection{Results on Downstream Classification Tasks}
Table \ref{tab:trans-model-performance} gives the transfer performance achieved by LaBSE on the SentEval benchmark~\cite{conneau-kiela-2018-senteval}, comparing against other state-of-the-art sentence embedding models. Despite its massive language coverage in a single model, LaBSE still obtains competitive transfer performance with monolingual English embedding models and the 16 language \emph{m}-USE model.

\begin{table}[!t]
\vskip -.1in
\small
\centering
    \resizebox{\linewidth}{!}{
    \begin{tabular}{l r r r r r r r}
    \hline
    {\bf Model} & {\bf MR} & {\bf CR} & {\bf SUBJ} & {\bf MPQA} & {\bf TREC} & {\bf SST} & {\bf MRPC} \\
    \hline
    \multicolumn{8}{c}{\textit{English Models}} \\
    InferSent                       & 81.1 & 86.3 & 92.4 & \textbf{90.2} & 88.2 & 84.6 & 76.2 \\
    Skip-Thought LN                 & 79.4 & 83.1 & 93.7 & 89.3 & -- & -- & -- \\
    Quick-Thought                   & \textbf{82.4} & 86.0 & 94.8 & \textbf{90.2} & 92.4 & \textbf{87.6} & \textbf{76.9} \\
    USE\textsubscript{Trans}        & 82.2 & 84.2 & \textbf{95.5} & 88.1 & 93.2 & 83.7 & -- \\
    \hline
    \multicolumn{8}{c}{\textit{Multilingual Models}} \\
    \emph{m}-USE\textsubscript{Trans}  & 78.1 & \textbf{87.0} & 92.1 & 89.9 & \textbf{96.6} & 80.9 & -- \\ 
    LaBSE                           & 79.1 & 86.7 & 93.6 & 89.6 & 92.6 & 83.8 & 74.4 \\
    \hline
    \end{tabular}}
\caption{Performance on English transfer tasks from SentEval \citep{conneau-kiela-2018-senteval}. We compare LaBSE model with InferSent~\cite{infersent}, Skip-Thought LN~\cite{ba2016}, Quick-Thought~\cite{quickthought2018}, USE\textsubscript{Trans}~\cite{use}, and \emph{m}-USE\textsubscript{Trans}~\cite{muse}.}
\label{tab:trans-model-performance}
\end{table}

\section{Analysis}

\subsection{Additive Margin Softmax}

\begin{figure}[!t]
  \centering
  \includegraphics[width=\linewidth]{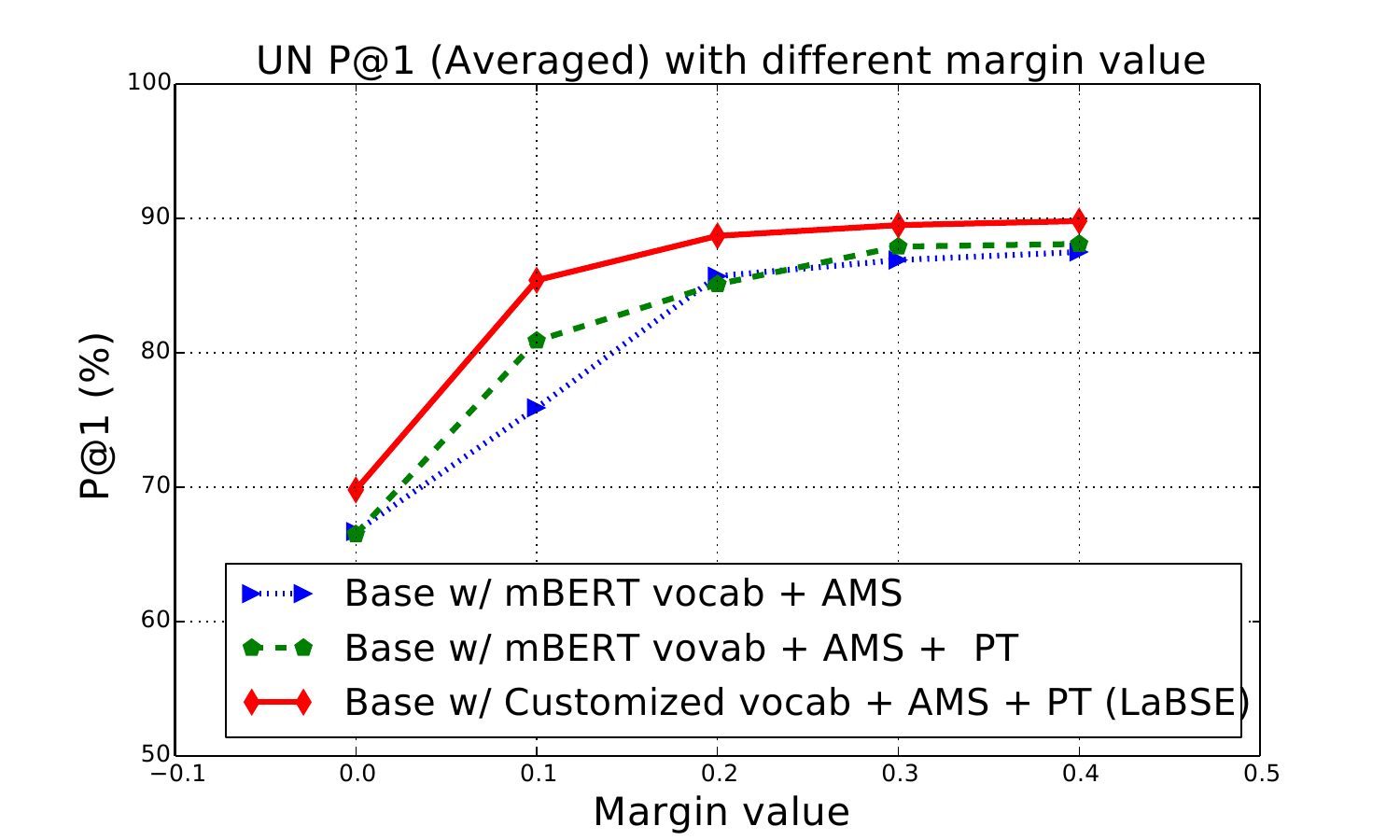}
  \vskip -.1in
  \caption{Average P@1~(\%) on UN retrieval task of models trained with different margin values.}
  \label{fig:un_margin}
  \vskip -.1in
\end{figure}

The above experiments show that additive margin softmax is a critical factor in learning good cross-lingual embeddings, which is aligned with the findings from \citet{yang2019improve}. 
We further investigate the effect of margin size on our three model variations, as shown in figure~\ref{fig:un_margin}.
The model with an additive margin value 0 performs poorly on the UN task with $\sim$60 average P@1 across all three model variations.
With a small margin value of 0.1, the model improves significantly compare to no margin with 70s to 80s average P@1.
Increasing the margin value keeps improving performance until it reaches 0.3.
The trend is consistent on all models.

\subsection{Effectiveness of Pre-training}
\label{sec:effec_pt}

\begin{figure}[!t]
  \centering
  \includegraphics[width=\linewidth]{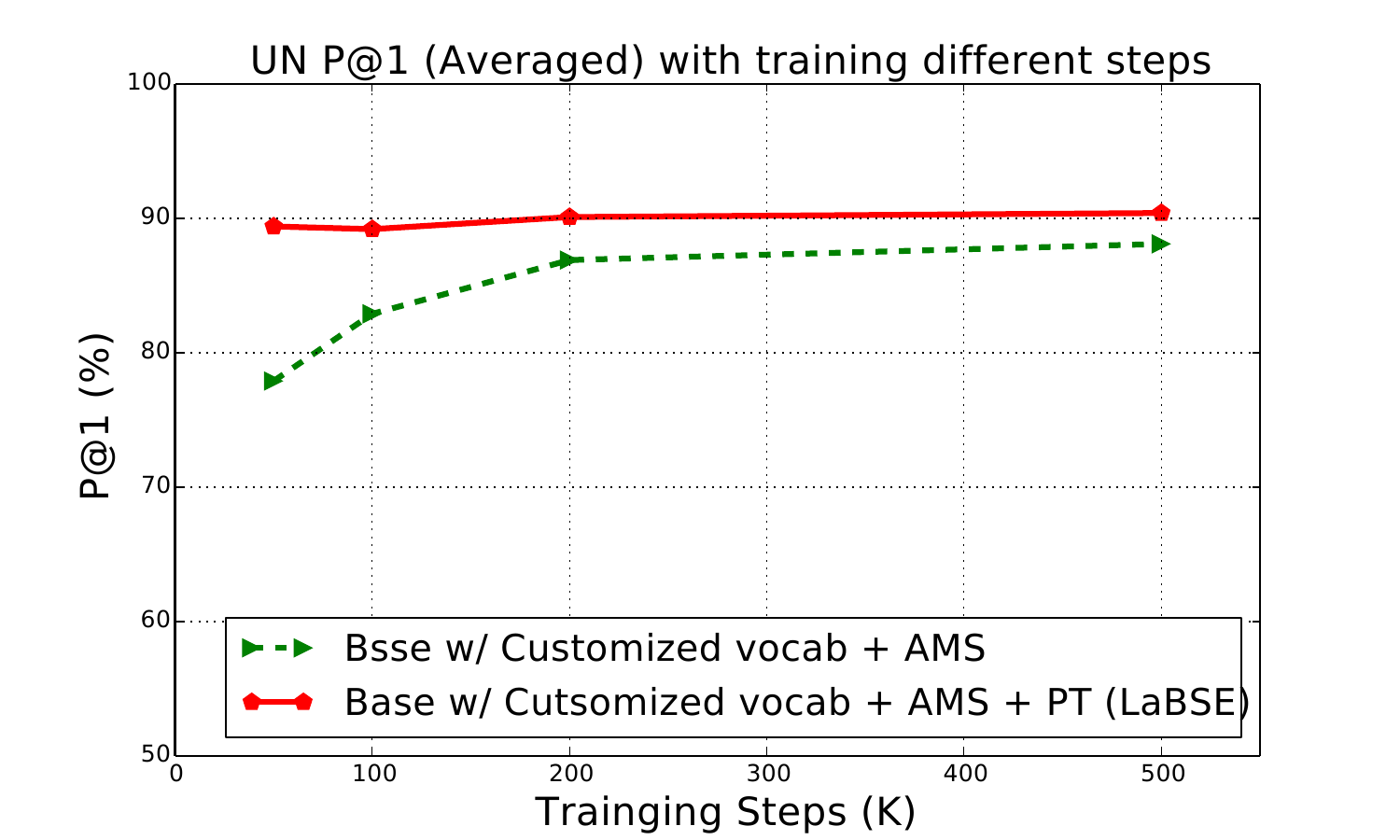}
  \vskip -.1in
  \caption{Average P@1~(\%) on UN retrieval task of models trained with training different steps.}
  \label{fig:un_steps}
  \vskip -.1in
\end{figure}

To better understand the effective of MLM/TLM pre-training in the final LaBSE model, we explore training a variant of this model using our customized vocab but without pre-training.
The results are shown in figure \ref{fig:un_steps}.
We experiment with varying the number of training steps for both models, including: 50k, 100K, 200K, and 500K steps.
A model with pre-trained encoders already achieves the highest performance when trained 50K steps, further training doesn't increase the performance significantly.
However, the model without pre-training performs poorly when only trained 50k steps.
Its performance increases with additional steps and approaches the model with pre-training at 500k steps.
The overall performance is, however, still slightly worse. Moreover, further training past 500k steps doesn't increase the performance significantly. Pre-training thus both improves performance and dramatically reduces the amount of parallel data required.
Critically, the model sees 1B examples at 500K steps, while the 50K model only sees 200M examples.\footnote{We note that it is relative easy to get 200M parallel examples for many languages from public sources like Paracrawl, TED58, while obtaining 1B examples is generally much more challenging. }

\subsection{Low Resource Languages and Languages without Explicit Training Data}
We evaluate performance through further experiments on Tatoeba for comparison to prior work and to identify broader trends.
Besides the 36 language group and all-languages group, 
two more groups of 14 languages (selected from the languages covered by \emph{m}-USE, and 82 languages group (covered by the LASER training data) are evaluated. 
Table \ref{tab:tatoeba} provides the macro-average accuracy achieved by LaBSE for the four language groupings drawn from Tatoeba, comparing against LASER and \emph{m}-USE.
All three models perform well on the 14 major languages support by \emph{m}-USE, with each model achieving an average accuracy $>$93\%.
Both LaBSE and LASER perform moderately better than \emph{m}-USE, with an accuracy of 95.3\%. As more languages are included, the averaged accuracy for both LaBSE and LASER decreases, but with a notably more rapid decline for LASER. LaBSE systematically outperforms LASER on the groups of 36 languages (+10.6\%), 82 languages (+11.4\%), and 112 languages (+18.2\%).

\begin{figure}[!t]
  \centering
  \includegraphics[width=\linewidth]{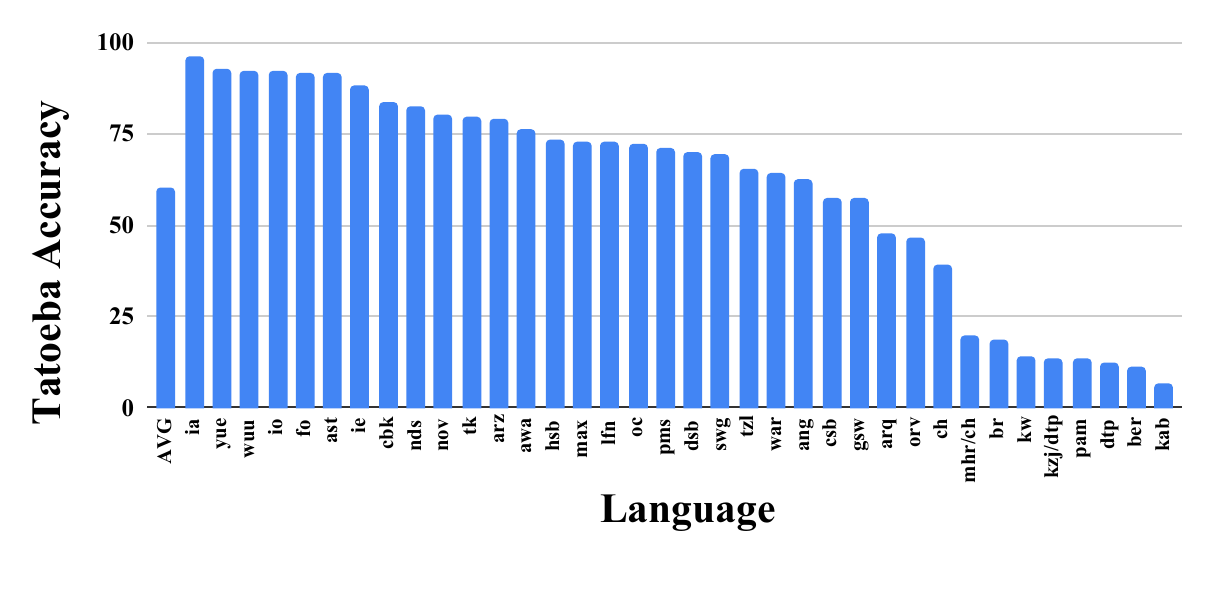}
  \vskip -.1in
  \caption{Tatoeba accuracy for those languages without any explicit training data. The average (AVG) accuracy is 60.5\%, listed at the first.}
  \label{fig:no_data}
  \vskip -.1in
\end{figure}

Figure \ref{fig:no_data} lists the Tatoeba accuracy for languages where we don't have any explicit training data. 
There are a total of 30+ such languages. 
The performance is surprisingly good for most of the languages with an average accuracy around 60\%.
Nearly one third of them have accuracy greater than 75\%, and only 7 of them have accuracy lower than 25\%. 
One possible reason is that language mapping is done manually and some languages are close to those languages with training data but may be treated differently according to ISO-639 standards and other information. Additional, since automatic language detection is used, some limited amount of data for the missing languages might be included during training.
We also suspect that those well performing languages are close to some language that we have training data.
For example \textit{yue} and \textit{wuu} are related to \textit{zh}~(Chinese) and \textit{fo} has similarities to \textit{is}~(ICELANDIC).
Multilingual generalization across so many languages is only possible due to the massively multilingual nature of LaBSE.

\begin{table}[!t]
    \centering
    \resizebox{\linewidth}{!}{
    \begin{tabular}{l | r r r r}
        \textbf{Model}  &\textbf{14~Langs} & \textbf{36~Langs} & \textbf{82~Langs} & \textbf{All~Langs} \\ \hline
         \emph{m}-USE\textsubscript{Trans.} & 93.9 & -- & -- & -- \\
         LASER  & \textbf{95.3} & 84.4 & 75.9 & 65.5 \\
         LaBSE  & \textbf{95.3} & \textbf{95.0} & \textbf{87.3} & \textbf{83.7} \\
    \end{tabular}
    }
    \caption{Accuracy(\%) of the Tatoeba datasets. \textbf{[14 Langs]}: The languages USE supports. \textbf{[36 Langs]}: The languages selected by XTREME. \textbf{[82 Langs]}: Languages that LASER has training data. \textbf{All Langs}: All languages supported by Taoteba.}
    \label{tab:tatoeba}
\end{table}

\subsection{Semantic Similarity}

The Semantic Textual Similarity (STS) benchmark~\cite{stsbenchmark} measures the ability of models to replicate fine-grained grained human judgements on pairwise English sentence similarity. Models are scored according to their Pearson correlation, $r$, on gold labels ranging from 0, unrelated meaning, to 5, semantically equivalent, with intermediate values capturing carefully defined degrees of meaning overlap. STS is used to evaluate the quality of sentence-level embeddings by assessing the degree to which similarity between pairs of sentence embeddings aligns with human perception of sentence meaning similarity.

Table \ref{tab:stsb} reports performance on the STS benchmark for LaBSE versus existing sentence embedding models. Following prior work, the semantic similarity of a sentence pair according to LaBSE is computed as the arc cosine distance between the pair's sentence embeddings.\footnote{Within prior work, \emph{m}-USE, USE and ConvEmbed use arccos distance to measure embedding space semantic similarity, while InferSent and SentenceBERT use cosine similarity.}
For comparison, we  include numbers for SentenceBERT when it is fine-tuned for the STS task as well as ConvEmbed when an additional affine transform is trained to fit the embeddings to STS. 
We observe that LaBSE performs worse on pairwise English semantic similarity than other sentence embedding models. We suspect training LaBSE on translation pairs biases the model to excel at detecting meaning equivalence, but not at distinguishing between fine grained degrees of meaning overlap.

Recently \citet{reimers-gurevych-2020-making} showed one can distill a English sentence representation model to a student multilingual model using a language alignment loss. The distilled model performs well on (multilingual-)STS benchmarks but underperforms on bitext retrieval tasks when compared to state-of-the-art models. Our approach is complimentary and can be combined with their method to distill better student models. 

\begin{table}[t]
     \centering
     \resizebox{\linewidth}{!}{
     \begin{tabular}{l|c c}
         \textbf{Model} &\textbf{dev} & \textbf{test}\\ \hline
         SentenceBERT \cite{SentenceBERT}  & - & 79.2 \\
         \emph{m}-USE \cite{muse}          &  \textbf{83.7} & \textbf{82.5} \\
         USE \cite{use}                    &  80.2 & 76.6 \\
         ConvEmbed \cite{yang-etal-2018-learning} & 81.4  & 78.2  \\
         InferSent \cite{infersent}        & 80.1 & 75.6 \\
         LaBSE                             & 74.3 & 72.8 \\
         \hline
         \multicolumn{2}{c}{STS Benchmark Tuned}\\
         \hline
         SentenceBERT-STS \cite{SentenceBERT} & - & 86.1 \\
         ConvEmbed \cite{yang-etal-2018-learning} & 83.5 & 80.8 \\
         \end{tabular}
     }
     \caption{Semantic Textual Similarity (STS) benchmark\cite{stsbenchmark} performance as measured by Pearson's $r$.}
     \label{tab:stsb}
     \vskip -.1in
\end{table}

\section{Mining Parallel Text from CommonCrawl}

We use the LaBSE model to mine parallel text from CommonCrawl, a large-scale multilingual web corpus, and then train NMT models on the mined data.
We experiment with two language pairs: English-to-Chinese (en-zh) and English-to-German (en-de).
We mine translations from monolingual CommonCrawl data processed as described above for self-supervised MLM pretraining.
After processing, there are 1.17B, 0.6B, 7.73B sentences for Chinese (zh), German (de), and English (en), respectively. LaBSE embeddings are used to pair each non-English sentence with its nearest English neighbor, dropping pairs with a similarity score $< 0.6$.\footnote{The threshold 0.6 is selected by manually inspecting a data sample, where pairs greater or equal to this threshold are likely to be translation or partial translation of each other.
This results in 715M and 302M  sentence pairs for en-zh and en-de, respectively. Note that the pairs may still be noisy, but we resort to  data selection to select sentence pairs in higher quality for training NMT models.}
For en-de and en-zh, we train a model with Transformer-Big~\cite{vaswani2017attention} in the following way: First we train the model on the mined data as is for 120k steps with batch size 10k. Then we select the best 20\% using \citet{wei2018}'s data selection method, and train for another 80k steps.

Results in table~\ref{tab:cc} show the effectiveness of the mined training data. By referencing previous results \cite{edunov-etal-2018-understanding}, we see the mined data yields performance that is only 2.8 BLEU away from performance of the best system that made use of the WMT17 en-de parallel data. Compare to prior en-zh results \cite{sennrich-etal-2017-university}, we see that the model is as good as a WMT17 NMT model \cite{sennrich-etal-2017-university} that is trained on the WMT en-zh parallel data. The table also gives BLEU performance on the TED test set~\cite{qi-etal-2018-pre}, with performance being comparable with models trained using CCMatrix~\cite{Schwenk2019CCMatrixMB}.\footnote{CCMatrix is another dataset contains billions of parallel sentences mined from CommonCrawl using a embedding based mining approach, with an additional cleaning step.} 

\begin{table}[!t]
    \centering
    \resizebox{\linewidth}{!}{
    \begin{tabular}{l|c c c c c}
        \multirow{2}{*}{\bf Langs}  & {\bf \# of}  & {\bf\# of} & {\bf\# of} &\multicolumn{2}{c}{\bf BLEU} \\
        & {\bf XX Sents} & {\bf En Sents} & {\bf Mined Pairs} & {\bf News} & {\bf TED} \\ \hline
        en-zh            & 1.17B & 7.73B & 715M & 36.3 & 15.2 \\
        en-de            &  601M & 7.73B & 302M & 28.1 & 31.3\\ 
    \end{tabular}
    }
    \caption{The number of source / target sentences and number of mined parallel text from CommonCrawl. BLEU scores~(en$\rightarrow$xx) are evaluated on WMT News dataset and TED dataset. We use wmtnews17 and wmtnews14 for zh-en and de-en respectively in WMT News set. }
    \label{tab:cc}
    \vskip -.1in
\end{table}

\section{Conclusion}
This paper presents a language-agnostic BERT sentence embedding (LaBSE) model supporting 109 languages.
The model achieves state-of-the-art performance on various bi-text retrieval/mining tasks compare to the previous state-of-the-art, while also providing increased language coverage.
We show the model performs strongly even on those languages where LaBSE doesn't have any explicit training data, likely due to language similarity and the massively multilingual natural of the model.
Extensive experiments show additive margin softmax is a key factor for training the model,
parallel data quality matters, but the effect of increased amounts of parallel data diminishes when a pre-trained language model is used.
The pre-trained model is released at \url{https://tfhub.dev/google/LaBSE}.

\section*{Acknowledgments}
We thank our teammates from Descartes, Translate and other Google groups for their feedback
and suggestions. Special thanks goes to Sidharth
Mudgal, and Jax Law for help with data processing; as well as Jialu Liu, Tianqi Liu, Chen Chen,
and Anosh Raj for help on BERT pretraining.

\bibliography{anthology,custom}
\bibliographystyle{acl_natbib}

\clearpage
\appendix

\section{LaBSE\textsubscript{Large}}
Motivated by the recent progress of giant models, we also train a model with increased model capacity. Following BERT\textsubscript{Large}, we develop LaBSE\textsubscript{Large} using a 24 layers transformer with 16 attention heads and 1024 hidden size. Constrained by computation resource, we train 1M steps one stage pre-training instead of the progressive multi-stage pre-training used when training LaBSE model. Fine-tuning configs are exact the same as the base LaBSE model. 

Table \ref{tab:large} shows the UN performance of the LaBSE\textsubscript{Large} model compared to LaBSE model. The results are mixed, and the average performances are very close. We also evaluate the model on Tatoeba, and the average performances across all languages are also very close: 83.7~(LaBSE) v.s. 83.8~(LaBSE\textsubscript{Large}).

\begin{table}[h!]
    \centering
    \resizebox{\linewidth}{!}{
    \begin{tabular}{l|c c c c c}
        \textbf{Model} &\textbf{es} & \textbf{fr} & \textbf{ru} & \textbf{zh} & \textbf{avg.} \\ \hline
        LaBSE                       & 91.1 & 88.3 & 90.8 & 87.7 & 89.5 \\
        LaBSE\textsubscript{Large}  & 90.9 & 87.9 & 89.4 & 89.5 & 89.4 \\
    \end{tabular}
    }
    \caption{P@1 on UN (en$\rightarrow$xx) .}
    \label{tab:large}
\end{table}

We suspect that the translate matching training objective is too easy, the model cannot learn more information from the current in-batch negative sampling approach. An improved negative contrast could help the larger model to learn better representations. We experimented with one type of hard negatives in the section below, but more types of hard negatives could be explored as described in \cite{jing2020neg}. We leave this as a future work.

\section{Hard Negative Mining}

Since their introduction into models that make use of dual encoders to learn cross-lingual embeddings, hard negatives~\cite{mandy2018} have become the de facto data augmentation method for learning cross-lingual sentence embeddings~\cite{muthu-crosslingual,yang2019improve}.
To get the hard negatives, a weaker dual encoder model is trained using a similar model but with less parameters and less training data.
For each training example, those incorrect translations that are semantically similar to the correct translation are retrieved as “hard-negatives” from a candidates pool. 
Semantically similarity is determined using the cosine similarity of the embeddings generated by the weaker model.
It is challenging to apply hard negative to large datasets as it is very time consuming and computationally costly .

We investigate hard negative mining closely following \citet{mandy2018}. 
By contacting the original authors, we obtained their negative mining pipeline, which employs a weaker dual encoder that uses a deep averaging network trained to identify translation pairs.
Similar to the cross-accelerator negatives, the mined negatives are also appended to each example. 

We only experiment using hard negative for Spanish~(es) as it is very costly to get hard negative for all languages.
Due to memory constraints, we only append 3 mined hard negatives in es for each en source sentence. 
Since the amount of examples increased 4x per en sentence in es batches,we also decrease batch size from 128 to 32 in the hard negative experiment. 
For languages other than es, the training data was the same as other the experiments but with batch size decreased to 32 together.
Other languages are trained as usual.
Table \ref{tab:negatives} shows the results of these models on UN.
The accuracy of all four languages went down, even for en-es where we have the hard negatives.
We suspect the worse performance is caused by the decreasing of batch size due to the memory constrain with more hard negative per example.

\begin{table}[h!]
    \centering
    \resizebox{\linewidth}{!}{
    \begin{tabular}{l|c c c c c}
        \textbf{Model} &\textbf{es} & \textbf{fr} & \textbf{ru} & \textbf{zh} & \textbf{avg.} \\ \hline
        LaBSE          & 91.1 & 88.3 & 90.8 & 87.7 & 89.5 \\
        LaBSE + es HN  & 90.4 & 87.1 & 89.9 & 87.2 & 88.7 \\
    \end{tabular}
    }
    \caption{P@1 on UN (en$\rightarrow$xx) with hard negative examples in en-es.}
    \label{tab:negatives}
\end{table}

\section{Supported Languages}
\begin{figure*}[!t]
  \centering
  \includegraphics[width=\linewidth]{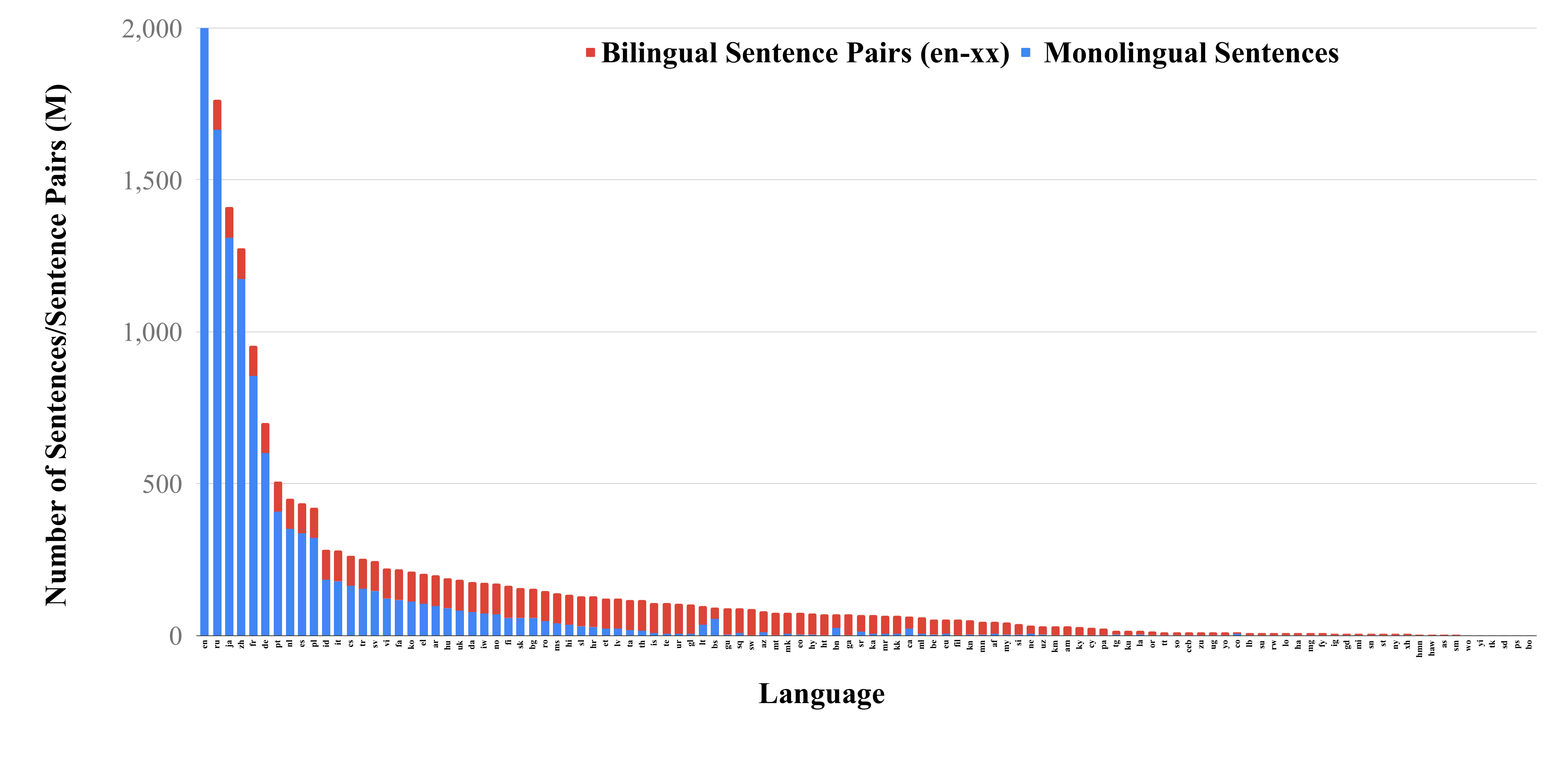}
  \caption{Quantity of monolingual sentences and bilingual sentence-pairs for each of the 109 languages in our training set. The English (en) sentences are capped at 2 billion.}
  \label{fig:data_count}
\end{figure*}

The supported langauges is listed in table \ref{tab:supported_langauges}.
The distribution for each supported language is shown in figure \ref{fig:data_count}.

\begin{table*}[!t]
    \centering
    \begin{tabular}{ll || ll || ll}
ISO & NAME & ISO & NAME  & ISO & NAME \\ \hline
af  & AFRIKAANS         & ht  & HAITIAN\_CREOLE   &  pt & PORTUGUESE      \\
am  & AMHARIC           & hu & HUNGARIAN          &  ro & ROMANIAN        \\
ar  & ARABIC            & hy & ARMENIAN           &  ru & RUSSIAN         \\
as  & ASSAMESE          & id & INDONESIAN         &  rw & KINYARWANDA     \\
az  & AZERBAIJANI       & ig & IGBO               &  si & SINHALESE       \\
be  & BELARUSIAN        & is & ICELANDIC          &  sk & SLOVAK          \\
bg  & BULGARIAN         & it & ITALIAN            &  sl & SLOVENIAN       \\
bn  & BENGALI           & ja & JAPANESE           &  sm & SAMOAN          \\
bo  & TIBETAN           & jv & JAVANESE           &  sn & SHONA           \\
bs  & BOSNIAN           & ka & GEORGIAN           &  so & SOMALI          \\
ca  & CATALAN           & kk & KAZAKH             &  sq & ALBANIAN        \\
ceb & CEBUANO           & km & KHMER              &  sr & SERBIAN         \\
co  & CORSICAN          & kn & KANNADA            &  st & SESOTHO         \\
cs  & CZECH             & ko & KOREAN             &  su & SUNDANESE       \\
cy  & WELSH             & ku & KURDISH            &  sv & SWEDISH         \\
da  & DANISH            & ky & KYRGYZ             &  sw & SWAHILI         \\
de  & GERMAN            & la & LATIN              &  ta & TAMIL           \\
el  & GREEK             & lb & LUXEMBOURGISH      &  te & TELUGU          \\
en  & ENGLISH           & lo & LAOTHIAN           &  tg & TAJIK           \\
eo  & ESPERANTO         & lt & LITHUANIAN         &  th & THAI            \\
es  & SPANISH           & lv & LATVIAN            &  tk & TURKMEN         \\
et  & ESTONIAN          & mg & MALAGASY           &  tl & TAGALOG         \\
eu  & BASQUE            & mi & MAORI              &  tr & TURKISH         \\
fa  & PERSIAN           & mk & MACEDONIAN         &  tt & TATAR           \\
fi  & FINNISH           & ml & MALAYALAM          &  ug & UIGHUR          \\
fr  & FRENCH            & mn & MONGOLIAN          &  uk & UKRAINIAN       \\
fy  & FRISIAN           & mr & MARATHI            &  ur & URDU            \\
ga  & IRISH             & ms & MALAY              &  uz & UZBEK           \\
gd  & SCOTS\_GAELIC     & mt & MALTESE            &  vi & VIETNAMESE      \\
gl  & GALICIAN          & my & BURMESE            &  wo & WOLOF           \\
gu  & GUJARATI          & ne & NEPALI             &  xh & XHOSA           \\
ha  & HAUSA             & nl & DUTCH              &  yi & YIDDISH         \\
haw & HAWAIIAN          & no & NORWEGIAN          &  yo & YORUBA          \\
he  & HEBREW            & ny & NYANJA             &  zh & CHINESE         \\
hi  & HINDI             & or & ORIYA              &  zu & ZULU            \\
hmn & HMONG             & pa & PUNJABI            &  & \\
hr  & CROATIAN          & pl & POLISH             &  & \\
    \end{tabular}
    \caption{The supported languages of LaBSE (ISO 639-1/639-2).}
    \label{tab:supported_langauges}
\end{table*}

\end{document}